\documentclass{article}

\usepackage{include/arxiv,times}

\usepackage{hyperref}
\usepackage{url}

\usepackage{algorithm}
\usepackage{algorithmicx}
\usepackage[noend]{algpseudocode}
\usepackage{changepage}

\algrenewcommand\algorithmicindent{1.0em}

\usepackage{graphicx}
\usepackage{comment}
\usepackage{amsmath}
\usepackage{amssymb}
\usepackage{siunitx}
\usepackage{wrapfig}
\usepackage{cleveref}

\usepackage{caption}
\usepackage{subcaption}

\usepackage{tipa}

\usepackage{pgffor}

\newcommand{\defeq}{:=}
\newcommand{\myvec}[1]{\mathbf{#1}}

\newcommand{\vb}{\myvec{b}}

\newcommand{\vh}{\myvec{h}}

\newcommand{\vr}{\myvec{r}}

\newcommand{\vu}{\myvec{u}}

\newcommand{\vw}{\myvec{w}}

\newcommand{\vx}{\myvec{x}}

\newcommand{\vz}{\myvec{z}}

\newcommand{\vU}{\myvec{U}}

\newcommand{\vW}{\myvec{W}}

\newcommand{\union}{\cup}

\newcommand{\be}{\begin{equation}}
\newcommand{\ee}{\end{equation}}
\newcommand{\bea}{\begin{eqnarray}}
\newcommand{\eea}{\end{eqnarray}}
\newcommand{\beaa}{\begin{eqnarray*}}
\newcommand{\eeaa}{\end{eqnarray*}}

\DeclareMathAlphabet{\mathpzc}{OT1}{pzc}{m}{n}
\DeclareMathOperator\sigm{sigm}

\newcommand{\minisec}[1]{\textbf{#1:}}

\title{LipNet: \hspace{-.5mm}End-\hspace{-.5mm}to-End \hspace{-.5mm}Sentence-level \hspace{-.5mm}Lipreading}

\author{Yannis M. Assael$^{1,\dagger}$, Brendan Shillingford$^{1,}$\thanks{These authors contributed equally to this work.}\ , Shimon Whiteson$^1$ \& Nando de Freitas$^{1,2,3}$\\
Department of Computer Science, University of Oxford, Oxford, UK $^1$\\
Google DeepMind, London, UK $^2$\\
CIFAR, Canada $^3$\\
\texttt{\{yannis.assael,brendan.shillingford,}\\\texttt{\ shimon.whiteson,nando.de.freitas\}@cs.ox.ac.uk}
}

\begin{document}

\maketitle
\begin{abstract}
	Lipreading is the task of decoding text from the movement of a speaker's mouth. 
Traditional approaches separated the problem into two stages: designing or learning visual features, and prediction. More recent deep lipreading approaches are end-to-end trainable \citep{wand2016lipreading,chung2016lip}.  
However, existing work on models trained end-to-end perform only word classification, rather than sentence-level sequence prediction.
Studies have shown that human lipreading performance increases for longer words \citep{easton1982perceptual}, indicating the importance of features capturing temporal context in an ambiguous communication channel. Motivated by this observation, we present LipNet, a model that maps a variable-length sequence of video frames to text, making use of spatiotemporal convolutions, a recurrent network, and the connectionist temporal classification loss, trained entirely end-to-end.
To the best of our knowledge, LipNet is the first end-to-end sentence-level lipreading model that simultaneously learns spatiotemporal visual features and a sequence model. On the GRID corpus, LipNet achieves $95.2\%$ accuracy in sentence-level, overlapped speaker split task, outperforming experienced human lipreaders and the previous $86.4\%$ word-level state-of-the-art accuracy \citep{gergen2016dynamic}.

\end{abstract}

\section{Introduction}
Lipreading plays a crucial role in human communication and speech understanding, as highlighted by the McGurk effect \citep{1976Nature}, where one phoneme's audio dubbed on top of a video of someone speaking a different phoneme results in a third phoneme being perceived.

Lipreading is a notoriously difficult task for humans, specially in the absence of context\footnote{LipNet video: 
{\fontsize{8}{8}\selectfont \url{https://youtube.com/playlist?list=PLXkuFIFnXUAPIrXKgtIpctv2NuSo7xw3k}}}.  
Most lipreading actuations, besides the lips and sometimes tongue and teeth, are latent and difficult to disambiguate without context \citep{fisher1968confusions,woodward1960phoneme}. For example, \citet{fisher1968confusions} gives 5 categories of visual phonemes (called \textit{visemes}), out of a list of 23 initial consonant phonemes, that are commonly confused by people when viewing a speaker's mouth. Many of these were asymmetrically confused, and observations were similar for final consonant phonemes.

Consequently, human lipreading performance is poor.  Hearing-impaired people achieve an accuracy of only $17\pm12\%$ even for a limited subset of 30 monosyllabic words and $21\pm11\%$ for 30 compound words \citep{easton1982perceptual}. 
An important goal, therefore, is to automate lipreading. Machine lipreaders have enormous practical potential, with applications in improved hearing aids, silent dictation in public spaces, security, speech recognition in noisy environments, biometric identification, and silent-movie processing.

Machine lipreading is difficult because it requires extracting spatiotemporal features from the video (since both position and motion are important). Recent deep learning approaches attempt to extract those features end-to-end. Most existing work, however, performs only word classification, not sentence-level sequence prediction.

In this paper, we present LipNet, which is to the best of our knowledge, the first \emph{end-to-end sentence-level} lipreading model.
As with modern deep learning based automatic speech recognition (ASR), LipNet is trained end-to-end to make sentence-level predictions. 
Our model operates at the character-level, using spatiotemporal convolutional neural networks (STCNNs), recurrent neural networks (RNNs), and the connectionist temporal classification loss (CTC)~\cite{graves2006connectionist}. 

Our empirical results on the GRID corpus \citep{cooke2006audio}, one of the few public sentence-level datasets, show that  LipNet attains a $95.2\%$ sentence-level word accuracy, in a overlapped speakers split that is popular for benchmarking lipreading methods. The previous best accuracy reported on an aligned word classification version of this task was $86.4\%$  \citep{gergen2016dynamic}. Furthermore, LipNet can generalise across unseen speakers in the GRID corpus with an accuracy of $88.6\%$.

We also compare the performance of LipNet with that of hearing-impaired people who can lipread on the GRID corpus task. On average, they achieve an accuracy of $52.3\%$, in contrast to LipNet's $1.69\times$ higher accuracy in the same sentences.

Finally, by applying saliency visualisation techniques \citep{zeiler2014visualizing,simonyan2013deep}, we interpret LipNet's learned behaviour, showing that the model attends to phonologically important regions in the video.
Furthermore, by computing intra-viseme and inter-viseme confusion matrices at the phoneme level, we show that almost all of LipNet's few erroneous predictions occur within visemes, since context is sometimes insufficient for disambiguation.

\vspace{-0.6ex}
\section{Related Work}

In this section, we outline various existing approaches to automated lipreading.

\minisec{Automated lipreading} Most existing work on lipreading does not employ deep learning.  Such work requires either heavy preprocessing of frames to extract image features, temporal preprocessing of frames to extract video features (e.g., optical flow or movement detection), or other types of handcrafted vision pipelines \citep{matthews2002extraction,zhao2009lipreading,gurban2009information,papandreou2007multimodal,papandreou2009adaptive,pitsikalis2006adaptive,lucey2006patch,papandreou2009adaptive}.
The automated lipreading literature is too vast to adequately cover, so we refer the reader to \citet{zhou2014review} for an extensive review.

Notably, \citet{goldschen1997continuous} were the first to do visual-only sentence-level lipreading using hidden Markov models (HMMs) in a limited dataset, using hand-segmented phones.
Later, \citet{neti2000audio} were the first to do sentence-level audiovisual speech recognition using an HMM combined with hand-engineered features, on the IBM ViaVoice~\citep{neti2000audio} dataset. The authors improve speech recognition performance in noisy environments by fusing visual features with audio ones.
The dataset contains 17111 utterances of 261 speakers for training (about 34.9 hours) and is not publicly available. 
As stated, their visual-only results cannot be interpreted as visual-only recognition, as they are used as rescoring of the noisy audio-only lattices. Using a similar approach, \citet{potamianos2003recent} report speaker independent and speaker adapted $91.62\%$, $82.31\%$ WER in the same dataset respectively, and $38.53\%$, $16.77\%$ WER in the connected DIGIT corpus, which contains sentences of digits.

Furthermore, \citet{gergen2016dynamic} use speaker-dependent training on an LDA-transformed version of the Discrete Cosine Transforms of the mouth regions in an HMM/GMM system. This work holds the previous state-of-the-art on the GRID corpus with a speaker-dependent accuracy of $86.4\%$.
Generalisation across speakers and extraction of motion features is considered an open problem, as noted in \citep{zhou2014review}.
LipNet addresses both of these issues.

\minisec{Classification with deep learning}
In recent years, there have been several attempts to apply deep learning to lipreading. However, all of these approaches perform only word or phoneme classification, whereas LipNet performs full sentence sequence prediction.
Approaches include learning multimodal audio-visual representations \citep{ngiam2011multimodal,sui2015listening,ninomiya2015integration,petridis2016deep},
learning visual features as part of a traditional speech-style processing pipeline (e.g. HMMs, GMM-HMMs, etc.) for classifying words and/or phonemes \citep{Almajai:2016,takashima2016audio,noda2014lipreading,koller2015deep},
or combinations thereof \citep{takashima2016audio}.
Many of these approaches mirror early progress in applying neural networks for acoustic processing in speech recognition \citep{hinton2012deep}.

\citet{chung2016lip} propose spatial and spatiotemporal convolutional neural networks, based on VGG, for word classification. The architectures are evaluated on a word-level dataset BBC TV (333 and 500 classes), but, as reported, their spatiotemporal models fall short of the spatial architectures by an average of around $14\%$. Additionally, their models cannot handle variable sequence lengths and they do not attempt sentence-level sequence prediction.

\citet{chung2016out} train an audio-visual max-margin matching model for learning pre-trained mouth features, which they use as inputs to an LSTM for 10-phrase classification on the OuluVS2 dataset, as well as a non-lipreading task.

\citet{wand2016lipreading} introduce LSTM recurrent neural networks for lipreading but address neither sentence-level sequence prediction nor speaker independence.

\citet{garglip} apply a VGG pre-trained on faces to classifying words and phrases from the MIRACL-VC1 dataset, which has only 10 words and 10 phrases.
However, their best recurrent model is trained by freezing the VGGNet parameters and then training the RNN, rather than training them jointly. Their best model achieves only $56.0\%$ word classification accuracy, and $44.5\%$ phrase classification accuracy, despite both of these being 10-class classification tasks.

\minisec{Sequence prediction in speech recognition}
The field of automatic speech recognition (ASR) would not be in the state it is today without modern advances in deep learning, many of which have occurred in the context of ASR \citep{graves2006connectionist,dahl2012context,hinton2012deep}.
The connectionist temporal classification loss (CTC) of \citet{graves2006connectionist}
drove the movement from deep learning as a component of ASR, to deep ASR systems trained end-to-end \citep{graves2014towards,maas2015lexicon,amodei2015deep}.
As mentioned earlier, much recent lipreading progress has mirrored early progress in ASR, but stopping short of sequence prediction.

LipNet is the first end-to-end model that performs sentence-level sequence prediction for visual speech recogntion. That is, we demonstrate the first work that takes as input as sequence of images and outputs a distribution over sequences of tokens; it is trained end-to-end using CTC and thus also does not require alignments.

\minisec{Lipreading Datasets}
Lipreading datasets (AVICar, AVLetters, AVLetters2, BBC TV, CUAVE, OuluVS1, OuluVS2) are plentiful \citep{zhou2014review,chung2016lip}, but most only contain single words or are too small.
One exception is the GRID corpus \citep{cooke2006audio}, which has audio and video recordings of $34$ speakers who produced $1000$ sentences each, for a total of $28$ hours across $34000$ sentences. \Cref{tbl:datasets} summarises state-of-the-art performance in each of the main lipreading datasets.

\begin{table}[htb]
    \caption{Existing lipreading datasets and the state-of-the-art accuracy reported on these. The size column represents the number of utterances used by the authors for training. Although the GRID corpus contains entire sentences, \cite{gergen2016dynamic} consider only the simpler case of predicting isolated words. LipNet predicts sequences and hence can exploit temporal context to attain much higher accuracy. Phrase-level approaches were treated as plain classification.}
\label{tbl:datasets}
\begin{center}
\begin{tabular}{llrlr}
\multicolumn{1}{c}{\bf Method}  &\multicolumn{1}{c}{\bf Dataset}  &\multicolumn{1}{c}{\bf Size}  &\multicolumn{1}{c}{\bf Output}  &\multicolumn{1}{c}{\bf Accuracy}
\\ \hline \\
\cite{fu2008classification} & AVICAR & $851$ & Digits & $37.9\%$ \\
\cite{hu2016temporal} & AVLetter & $78$ & Alphabet & $64.6\%$ \\
\cite{papandreou2009adaptive} & CUAVE & $1800$ & Digits & $83.0\%$ \\
\cite{chung2016lip} & OuluVS1 & $200$ & Phrases & $91.4\%$ \\
\cite{chung2016out} & OuluVS2 & $520$ & Phrases & $94.1\%$ \\
\cite{chung2016lip} & BBC TV & $>400000$ & Words & $65.4\%$ \\
\cite{gergen2016dynamic} & GRID & $29700$ & Words$^\star$ & $86.4\%$ \\
LipNet & GRID & $28775$ & \textbf{Sentences} & $\mathbf{95.2\%}$ \\
\end{tabular}
\end{center}
\end{table}

We use the GRID corpus to evaluate LipNet because it is sentence-level and has the most data. The sentences are drawn from the following simple grammar:
$command^{(4)} + color^{(4)} + preposition^{(4)} + letter^{(25)} + digit^{(10)} + adverb^{(4)},$ where the number denotes how many word choices there are for each of the 6 word categories. The categories consist of, respectively,
$\{$bin, lay, place, set$\}$, $\{$blue, green, red, white$\}$, $\{$at, by, in, with$\}$, $\{A,\dots, Z\} \backslash \{W\}$, $\{$zero, \dots, nine$\}$, and $\{$again, now, please, soon$\}$, yielding $64000$ possible sentences.
For example, two sentences in the data are ``set blue by A four please'' and ``place red at C zero again''.

\section{LipNet}
\label{sec:model}

LipNet is a neural network architecture for lipreading that maps variable-length sequences of video frames to text sequences, and is trained end-to-end.  In this section, we describe LipNet's building blocks and architecture.

\subsection{Spatiotemporal Convolutions}

Convolutional neural networks (CNNs), containing stacked convolutions operating spatially over an image, have been instrumental in advancing performance in computer visions tasks such as object recognition that receive an image as input \citep{krizhevsky2012imagenet}. A basic 2D convolution layer from $C$ channels to $C'$ channels (without a bias and with unit stride) computes
\[
    [\mathrm{conv}(\vx,\vw)]_{c'ij} = \sum_{c=1}^C \sum_{i'=1}^{k_w} \sum_{j'=1}^{k_h} w_{c'ci'j'} x_{c,i+i',j+j'}, 
\]
for input $\vx$ and weights $\vw\in\mathbb{R}^{C'\times C\times k_w\times k_h}$ where we define $x_{cij} = 0$ for $i,j$ out of bounds.
Spatiotemporal convolutional neural networks (STCNNs) can process video data by convolving across time, as well as the spatial dimensions \citep{karpathy2014large,ji20133d}. Hence similarly,
\[
    [\mathrm{stconv}(\vx,\vw)]_{c'tij} = \sum_{c=1}^C \sum_{t'=1}^{k_t} \sum_{i'=1}^{k_w} \sum_{j'=1}^{k_h} w_{c'ct'i'j'} x_{c,t+t',i+i',j+j'}.
\]

\subsection{Gated Recurrent Unit}
\label{sec:lstm}

Gated Recurrent Unit (GRU) \citep{chung2014empirical} is a type of recurrent neural network (RNN) that improves upon earlier RNNs by adding cells and gates for propagating information over more time-steps and learning to control this information flow. It is similar to the Long Short-Term Memory (LSTM) RNN \citep{hochreiter1997long}.
We use the standard formulation:
\begin{align*}
    [\vu_t, \vr_t]^T &= \sigm(\vW_z \vz_t + \vW_h \vh_{t-1} + \vb_g ) \\
    \tilde\vh_t &= \tanh(\vU_z \vz_t + \vU_h (\vr_t \odot \vh_{t-1}) + \vb_h ) \\
    \vh_t &= (\mathbf 1 - \vu_t) \odot \vh_{t-1} + \vu_t \odot \tilde\vh_t
\end{align*}
where $\vz\defeq\{\vz_1, \ldots, \vz_T\}$ is the input sequence to the RNN, $\odot$ denotes element-wise multiplication, and $\sigm(r) = 1/(1+\exp(-r))$.
We use a bidirectional GRU (Bi-GRU) as introduced by \cite{graves2005framewise} in the context of LSTMs: one RNN maps $\{\vz_1,\ldots,\vz_T\} \mapsto \{\overrightarrow{\vh_1}, \ldots, \overrightarrow{\vh_T}\}$, and another $\{\vz_T,\ldots,\vz_1\} \mapsto \{\overleftarrow{\vh_1}, \ldots, \overleftarrow{\vh_T}\}$, then $\vh_t\defeq[\overrightarrow{\vh_t}, \overleftarrow{\vh_t}]$. The Bi-GRU ensures that $\vh_t$ depends on $\vz_{t'}$ for all $t'$.
To parameterise a distribution over sequences, at time-step $t$ let $p(u_t|\vz) = \mathrm{softmax}(\mathrm{mlp}(\vh_t; \vW_{mlp}))$, where $\mathrm{mlp}$ is a feed-forward network with weights $\vW_{mlp}$. Then we can define the distribution over length-$T$ sequences as $p(u_1,\ldots,u_T | \vz) = \prod_{1\leq t\leq T} p(u_t | \vz)$, where $T$ is determined by $\vz$, the input to the GRU. In LipNet, $\vz$ is the output of the STCNN.

\subsection{Connectionist Temporal Classification}
\label{sec:ctc}

The connectionist temporal classification (CTC) loss \citep{graves2006connectionist} is widely used in modern speech recognition as it eliminates the need for training data that aligns inputs to target outputs \citep{amodei2015deep,graves2014towards,maas2015lexicon}. Given a model that outputs a sequence of discrete distributions over the token classes (vocabulary) augmented with a special ``blank'' token, CTC computes the probability of a sequence by marginalising over all sequences that are defined as equivalent to this sequence. This simultaneously removes the need for alignments and addresses variable-length sequences. 
Let $V$ denote the set of tokens that the model classifies at a single time-step of its output (vocabulary), and the blank-augmented vocabulary $\tilde V=V\union\{\textvisiblespace\}$ where \textvisiblespace{}~denotes the CTC blank symbol. Define the function $\mathcal B : \tilde V^* \to V^*$ that, given a string over $\tilde V$, deletes adjacent duplicate characters and removes blank tokens. For a label sequence $y\in V^*$, CTC defines $p(y|\vx) = \sum_{u\in \mathcal B^{-1}(y)\text{ s.t. }|u| = T} p(u_1,\ldots,u_T|\vx)$, where $T$ is the number of time-steps in the sequence model.
For example, if $T=3$, CTC defines the probability of a string ``$am$'' as $p(aam)+p(amm)+p(\textvisiblespace{}am)+p(a\textvisiblespace{}m)+p(am\textvisiblespace)$.
This sum is computed efficiently by dynamic programming, allowing us to perform maximum likelihood.

\subsection{LipNet Architecture}
\begin{figure}[tb]
  \centering
  \includegraphics[width=1.0\linewidth]{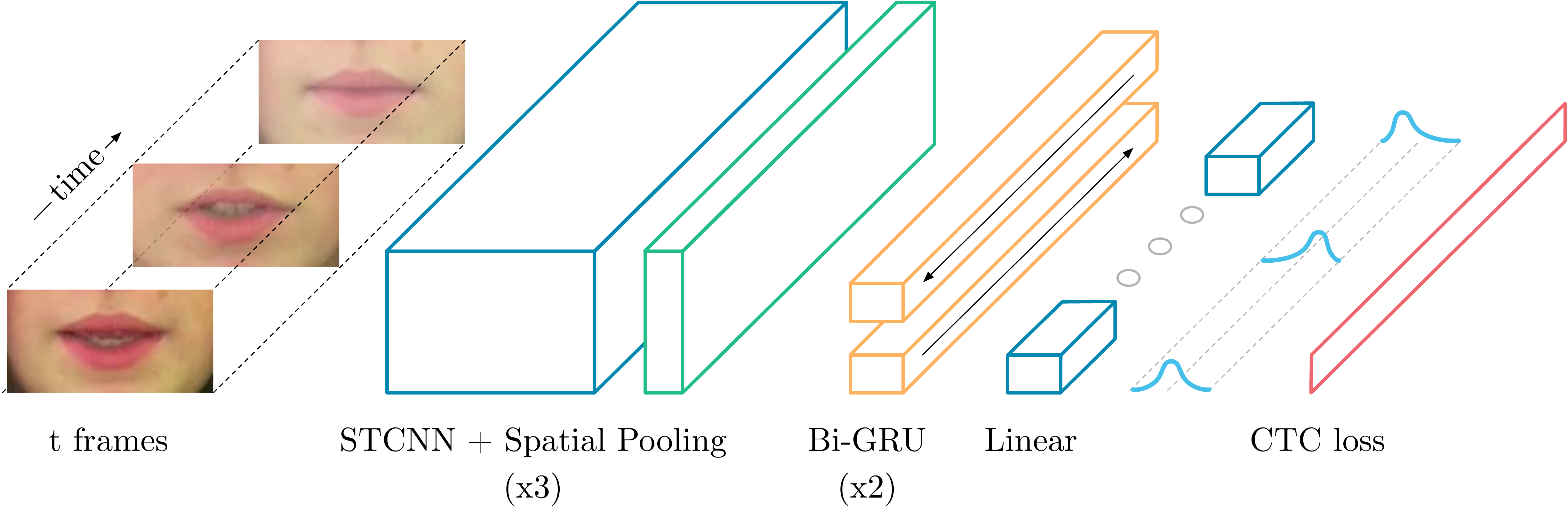}
  \vspace{-1.5em}
  \caption{LipNet architecture. A sequence of $T$ frames is used as input, and is processed by 3 layers of STCNN, each followed by a spatial max-pooling layer. The features extracted are processed by 2 Bi-GRUs; each time-step of the GRU output is processed by a linear layer and a softmax. This end-to-end model is trained with CTC.}  \label{fig:net-arch}
    \vspace{-1.em}
\end{figure}

\Cref{fig:net-arch} illustrates the LipNet architecture, which starts with $3\times$(spatiotemporal convolutions, channel-wise dropout, spatial max-pooling).
Subsequently, the features extracted are followed by two Bi-GRUs. The Bi-GRUs are crucial for efficient further aggregation of the STCNN output.
Finally, a linear transformation is applied at each time-step, followed by a softmax over the vocabulary augmented with the CTC blank, and then the CTC loss. All layers use rectified linear unit (ReLU) activation functions. More details including hyperparameters can be found in \Cref{tbl:arch} of Appendix~\ref{app:hyperparams}.

\newcommand\speakerindep{Unseen Speakers }
\newcommand\speakerdep{Overlapped Speakers }

\section{Lipreading Evaluation}

In this section, we evaluate LipNet on the GRID corpus. The augmentation methods employed don't make use of external data and rely purely on the GRID corpus.

\subsection{Data Augmentation}

\minisec{Preprocessing} The GRID corpus consists of 34 subjects, each narrating $1000$ sentences. The videos for speaker 21 are missing, and a few others are empty or corrupt, leaving $32746$ usable videos.
We employ a split (unseen speakers; not previously used in the literature) holding out the data of two male speakers (1 and 2) and two female speakers (20 and 22) for evaluation ($3971$ videos). The remainder is used for training ($28775$ videos).
We also use a sentence-level variant of the split (overlapped speakers) similar to \cite{wand2016lipreading}, where 255 random sentences from each speaker are used for evaluation. All remaining data from all speakers is pooled together for training.
All videos are 3 seconds long with a frame rate of 25fps. The videos were processed with the DLib face detector, and the iBug face landmark predictor~\citep{sagonas2013300} with 68 landmarks coupled with an online Kalman Filter. Using these landmarks, we apply an affine transformation to extract a mouth-centred crop of size $100\times50$ pixels per frame. We standardise the RGB channels over the whole training set to have zero mean and unit variance.

\minisec{Augmentation} We augment the dataset with simple transformations to reduce overfitting. First, we train on both the regular and the horizontally mirrored image sequence. Second, since the dataset provides word start and end timings for each sentence video, we augment the sentence-level training data with video clips of individual words as additional training instances. These instances have a decay rate of $0.925$. Third, to encourage resilience to varying motion speeds by deletion and duplication of frames, this is performed with a per-frame probability of $0.05$. The same augmentation methods were followed in all proposed baselines and models.

\subsection{Baselines}

To evaluate LipNet, we compare its performance to that of three hearing-impaired people who can lipread, as well as three ablation models inspired by recent state-of-the-art work \citep{chung2016lip,wand2016lipreading}.

\minisec{Hearing-Impaired People} This baseline was performed by three members of the Oxford Students' Disability Community. After being introduced to the grammar of the GRID corpus, they observed $10$ minutes of annotated videos from the training dataset, then annotated $300$ random videos from the evaluation dataset. When uncertain, they were asked to pick the most probable answer.

\minisec{Baseline-LSTM} Using the sentence-level training setup of LipNet, we replicate the model architecture of the previous deep learning GRID corpus state-of-the-art \citep{wand2016lipreading}. See Appendix~\ref{app:hyperparams} for more implementation details.

\minisec{Baseline-2D} Based on the LipNet architecture, we replace the STCNN with spatial-only convolutions similar to those of \cite{chung2016lip}. Notably, contrary to the results we observe with LipNet, \cite{chung2016lip} report $14\%$ and $31\%$ poorer performance of their STCNNs compared to the 2D architectures in their two datasets.

\minisec{Baseline-NoLM} Identical to LipNet, but with the language model used in beam search disabled.

\subsection{Performance Evaluation}
To measure the performance of LipNet and the baselines, we compute the word error rate (WER) and the character error rate (CER), standard metrics for the performance of ASR models. We produce approximate maximum-probability predictions from LipNet by performing CTC beam search. WER (or CER) is defined as the minimum number of word (or character) insertions, substitutions, and deletions required to transform the prediction into the ground truth, divided by the number of words (or characters) in the ground truth. Note that WER is usually equal to classification error when the predicted sentence has the same number of words as the ground truth, particularly in our case since almost all errors are substitution errors.

\Cref{tbl:perf} summarises the performance of LipNet compared to the baselines. According to the literature, the accuracy of human lipreaders is around $20\%$ \citep{easton1982perceptual,hilder2009comparison}. As expected,  the fixed sentence structure and the limited subset of words for each position in the GRID corpus facilitate the use of context, increasing performance. On the unseen speakers split, the three hearing-impaired people  achieve $57.3\%$, $50.4\%$, and $35.5\%$ WER respectively, yielding an average of $47.7\%$ WER.

\begin{table}[htb]
\caption{Performance of LipNet on the GRID dataset compared to the baselines, measured on two splits: (a) evaluating on only unseen speakers, and (b) evaluating on a 255 video subset of each speakers' sentences.}
\label{tbl:perf}
\begin{center}
\begin{tabular}{l|rr|rr}
\multicolumn{1}{c}{}  &\multicolumn{2}{c}{\bf\speakerindep}  &\multicolumn{2}{c}{\bf\speakerdep} \\
\multicolumn{1}{c}{\bf Method}  &\multicolumn{1}{c}{\bf CER}  &\multicolumn{1}{c}{\bf WER} &\multicolumn{1}{c}{\bf CER}  &\multicolumn{1}{c}{\bf WER}
\\ \hline\vspace{-.7em} \\ 
Hearing-Impaired Person (avg)
	& $-$ & $47.7\%$
	& $-$ & $-$ \\ 
Baseline-LSTM
	& $38.4\%$ & $52.8\%$
    & $15.2\%$ & $26.3\%$\\
Baseline-2D
	& $16.2\%$ & $26.7\%$
    & $4.3\%$ & $11.6\%$\\
Baseline-NoLM
	& $6.7\%$ & $13.6\%$
    & $2.0\%$ & $5.6\%$\\
LipNet
    & $\mathbf{6.4}\%$ & $\mathbf{11.4}\%$
    & $\mathbf{1.9\%}$ & $\mathbf{4.8\%}$ 
\end{tabular} 
\end{center}
\end{table}

For both unseen and overlapped speakers evaluation, the highest performance is achieved by the architectures enhanced with convolutional stacks. 
LipNet exhibits a $2.3\times$ higher performance in the overlapped compared to the unseen speakers split.
For unseen speakers, Baseline-2D and LipNet achieve $1.8\times$ and $4.2\times$ lower WER, respectively, than hearing-impaired people. 

The WER for unseen speakers Baseline-2D is $26.7\%$, whereas for LipNet it is $2.3\times$ lower, at $11.4\%$. Similarly, the error rate for overlapped speakers was $2.4\times$ lower for LipNet compared to Baseline-2D. Both results demonstrate the importance of combining STCNNs with RNNs.
This performance difference confirms the intuition that extracting spatiotemporal features using a STCNN is better than aggregating spatial-only features. 
This observation contrasts with the empirical observations of \cite{chung2016lip}.
Furthermore, LipNet's use of STCNN, RNNs, and CTC cleanly allow processing both variable-length input and variable-length output sequences, whereas the architectures of \citet{chung2016lip} and \citet{chung2016out} only handle the former.

Baseline-LSTM exhibits the lowest performance, in both unseen and overlapped speakers, with $52.8\%$ and $26.3\%$ WER, respectively. 
Interestingly, although Baseline-LSTM replicates the architecture of \citet{wand2016lipreading}, and despite the numerous data augmentation methods, the model performs $1.3\times$ lower than the reported $79.6\%$ word-level accuracy illustrating the difficulty of a sentence-level task even in a restricted grammar.

Finally, by disabling the language model, the Baseline-NoLM exhibits approximately $1.2\times$ higher WER than our proposed model.

\subsection{Learned representations}

In this section, we analyse the learned representations of LipNet from a phonological perspective. First, we create saliency visualisations \citep{simonyan2013deep,zeiler2014visualizing} to illustrate where LipNet has learned to attend. In particular, we feed an input into the model and greedily decode an output sequence, yielding a CTC alignment $\hat u\in \tilde V^*$ (following the notation of Sections~\ref{sec:lstm} and~\ref{sec:ctc}). Then, we compute the gradient of $\sum_t p(\hat u_t|\vx)$ with respect to the input video frame sequence, but unlike \citet{simonyan2013deep}, we use guided backpropagation \citep{springenberg2014striving}. Second, we train LipNet to predict ARPAbet phonemes, instead of characters, to analyse visual phoneme similarities using intra-viseme and inter-viseme confusion matrices.

\subsubsection{Saliency Maps}

\begin{figure}[htb]
	\vspace{-0.5em}
    \centering
    \includegraphics[width=1.0\linewidth]{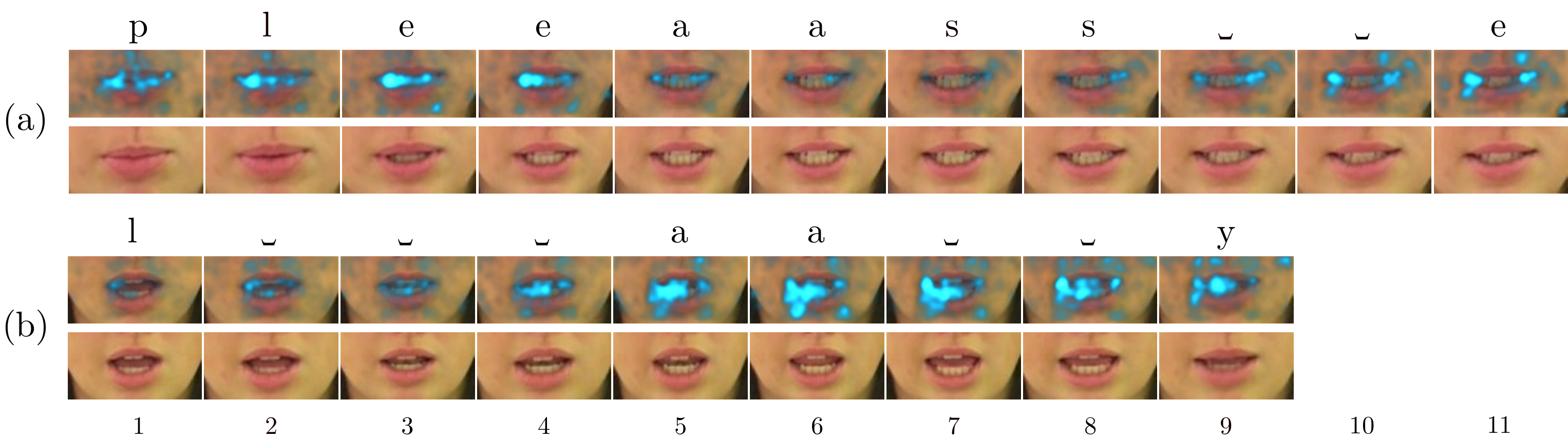}
    \vspace{-1.6em}
    \caption{Saliency maps for the words (a) \textit{please} and (b) \textit{lay}, produced by backpropagation to the input, showing the places where LipNet has learned to attend. The pictured transcription is given by greedy CTC decoding. CTC blanks are denoted by `\textvisiblespace'.}
    \label{fig:saliency}
	\vspace{-0.5em}
\end{figure}

We apply saliency visualisation techniques to interpret LipNet's learned behaviour, showing that the model attends to phonologically important regions in the video. In particular, in~\Cref{fig:saliency} we analyse two saliency visualisations for the words \textit{please} and \textit{lay} for speaker~25, based on \cite{ashby2013understanding}.  

The production of the word \textit{please} requires a great deal of articulatory movement at the beginning: the lips are pressed firmly together for the bilabial plosive /p/ (frame~1). At the same time, the blade of the tongue comes in contact with the alveolar ridge in anticipation of the following lateral /l/. The lips then part, allowing the compressed air to escape between the lips (frame~2). The jaw and lips then open further, seen in the distance between the midpoints of the upper and lower lips, and the lips spread (increasing the distance between the corners of the mouth), for the close vowel /iy/ (frame~3--4).  Since this is a relatively steady-state vowel,  lip position remains unchanged for the rest of its duration (frames~4--8), where the attention level drops considerably. The jaw and the lips then close slightly, as the blade of the tongue needs to be brought close to the alveolar ridge, for /z/ (frames~9--10), where attention resumes. 

\textit{Lay} is interesting since the bulk of frontally visible articulatory movement involves the blade of the tongue coming into contact with the alveolar ridge for /l/ (frames~2--6), and then going down for the vowel /ey/ (frames~7--9). That is exactly where most of LipNet's attention is focused, as there is little change in lip position. 

\subsubsection{Visemes}

According to \citet{deland1931story} and \citet{fisher1968confusions}, Alexander Graham Bell first hypothesised that multiple phonemes may be visually identical on a given speaker. This was later verified, giving rise to the concept of a \emph{viseme}, a visual equivalent of a phoneme \citep{woodward1960phoneme,fisher1968confusions}.
For our analysis, we use the phoneme-to-viseme mapping of \cite{neti2000audio}, clustering the phonemes into the following categories: Lip-rounding based vowels (V), Alveolar-semivowels (A), Alveolar-fricatives (B), Alveolar (C), Palato-alveolar (D), Bilabial (E), Dental (F), Labio-dental (G), and Velar (H). The full mapping can be found in \Cref{tbl:neti} in Appendix~\ref{app:hyperparams}. The GRID corpus contain 31 out of the 39 phonemes in ARPAbet.
\begin{figure}[htb]
    \centering
    \begin{subfigure}[t]{0.22\textwidth}
        \centering
        \includegraphics[width=1\linewidth]{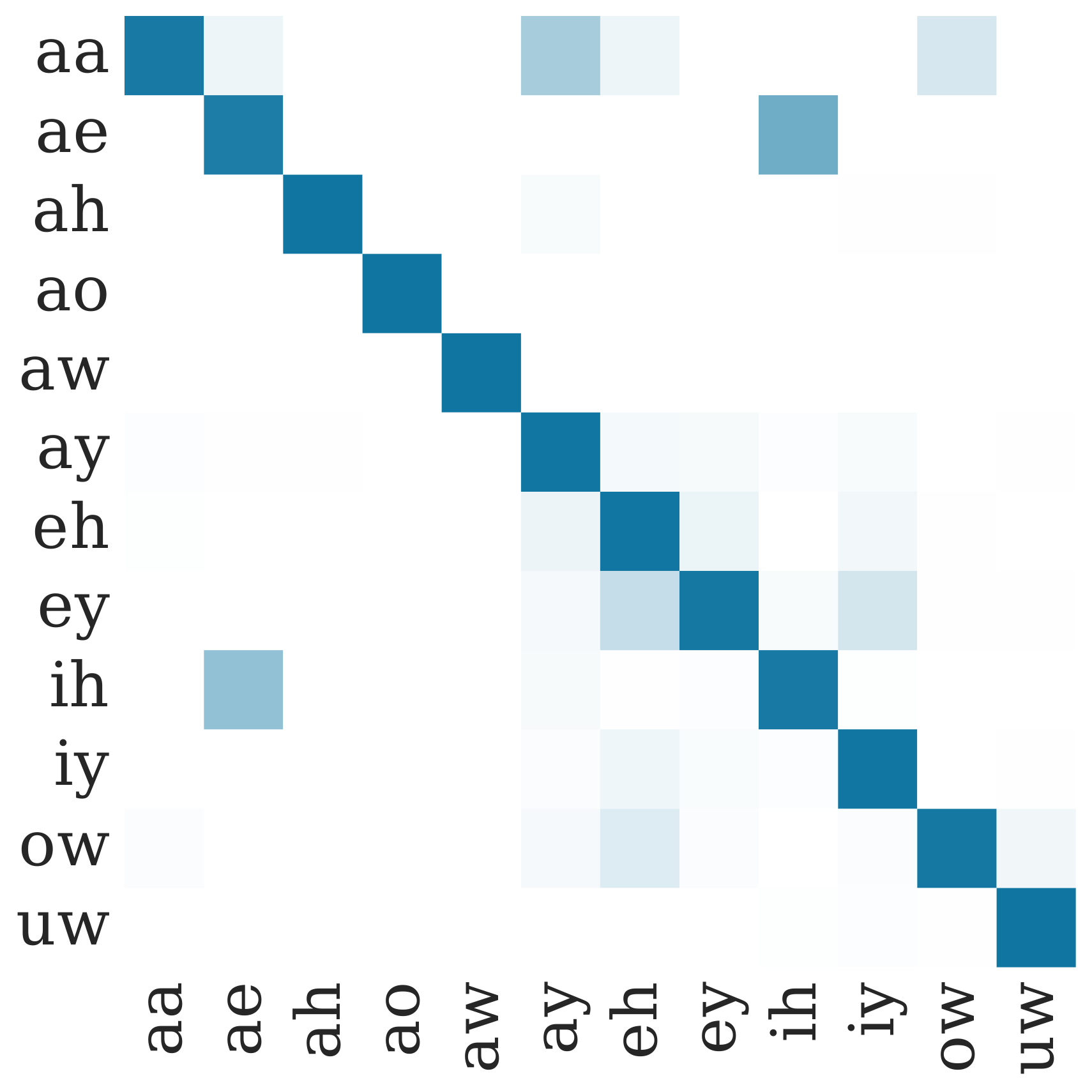}
        \caption{Lip-rounding vowels}
    \end{subfigure}%
    ~
    \begin{subfigure}[t]{0.22\textwidth}
        \centering
        \includegraphics[width=1\linewidth]{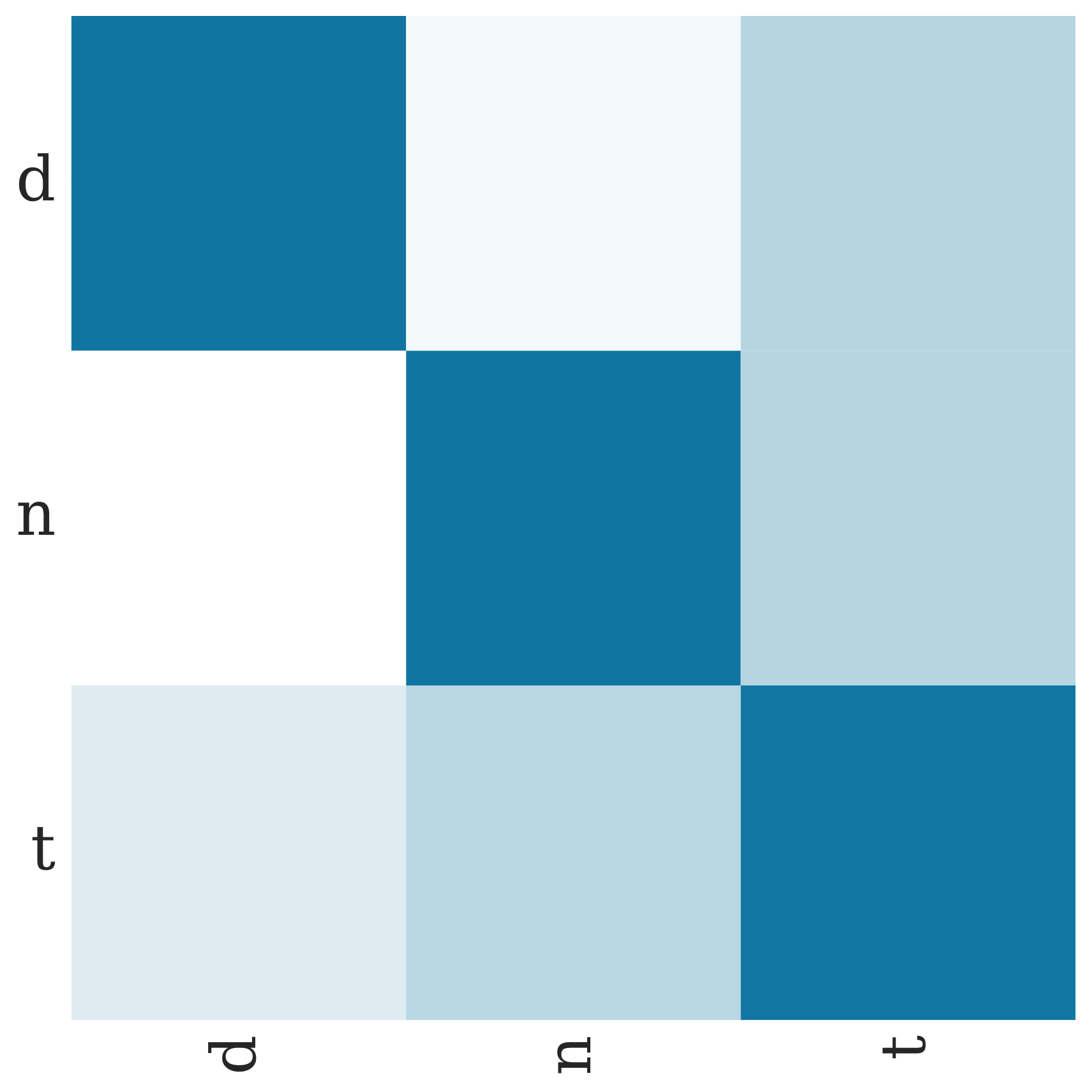}
        \caption{Alveolar}
    \end{subfigure}%
    ~
    \begin{subfigure}[t]{0.22\textwidth}
        \centering
        \includegraphics[width=1\linewidth]{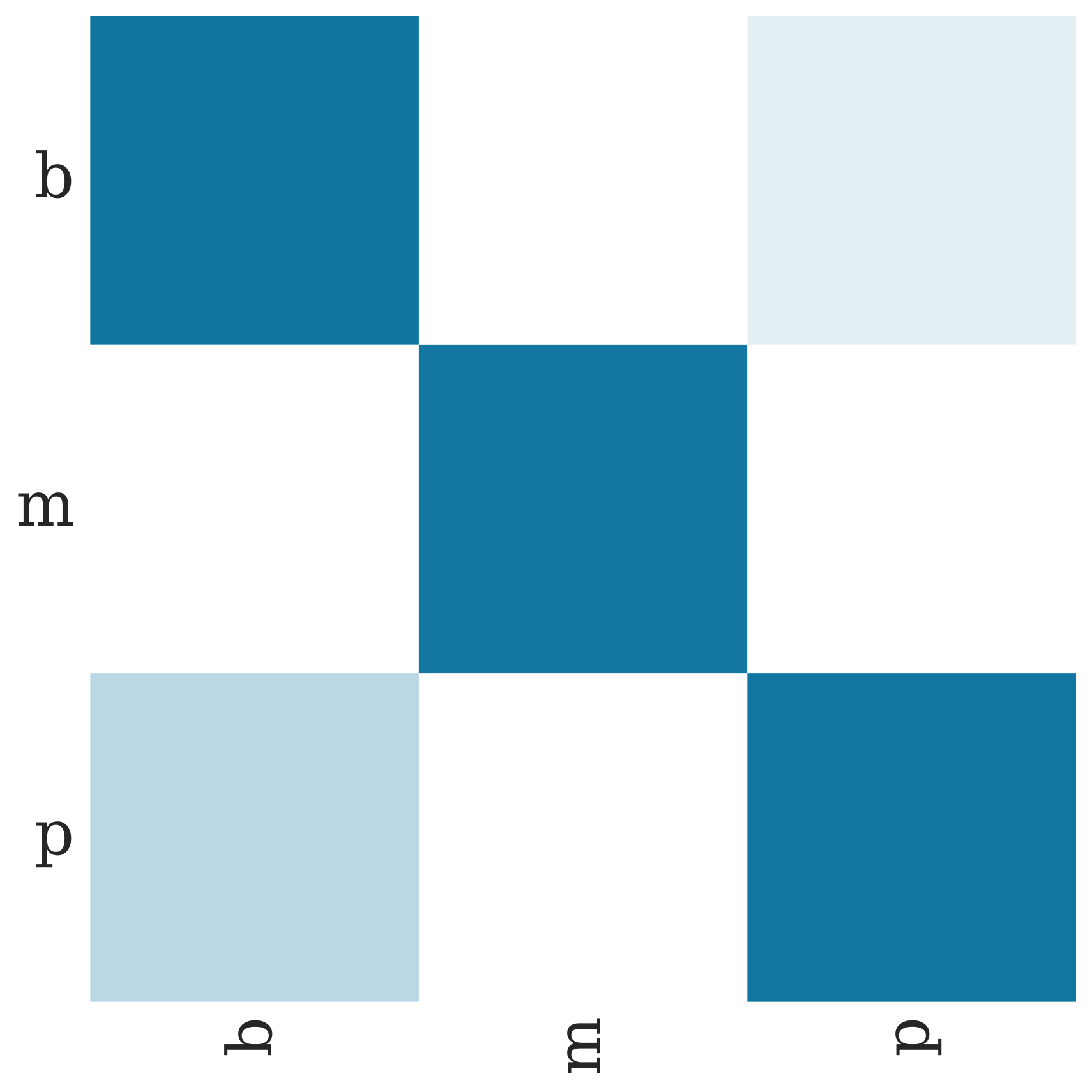}
        \caption{Bilabial}
    \end{subfigure}%
    ~
    \begin{subfigure}[t]{0.22\textwidth}
        \centering
        \includegraphics[width=1\linewidth]{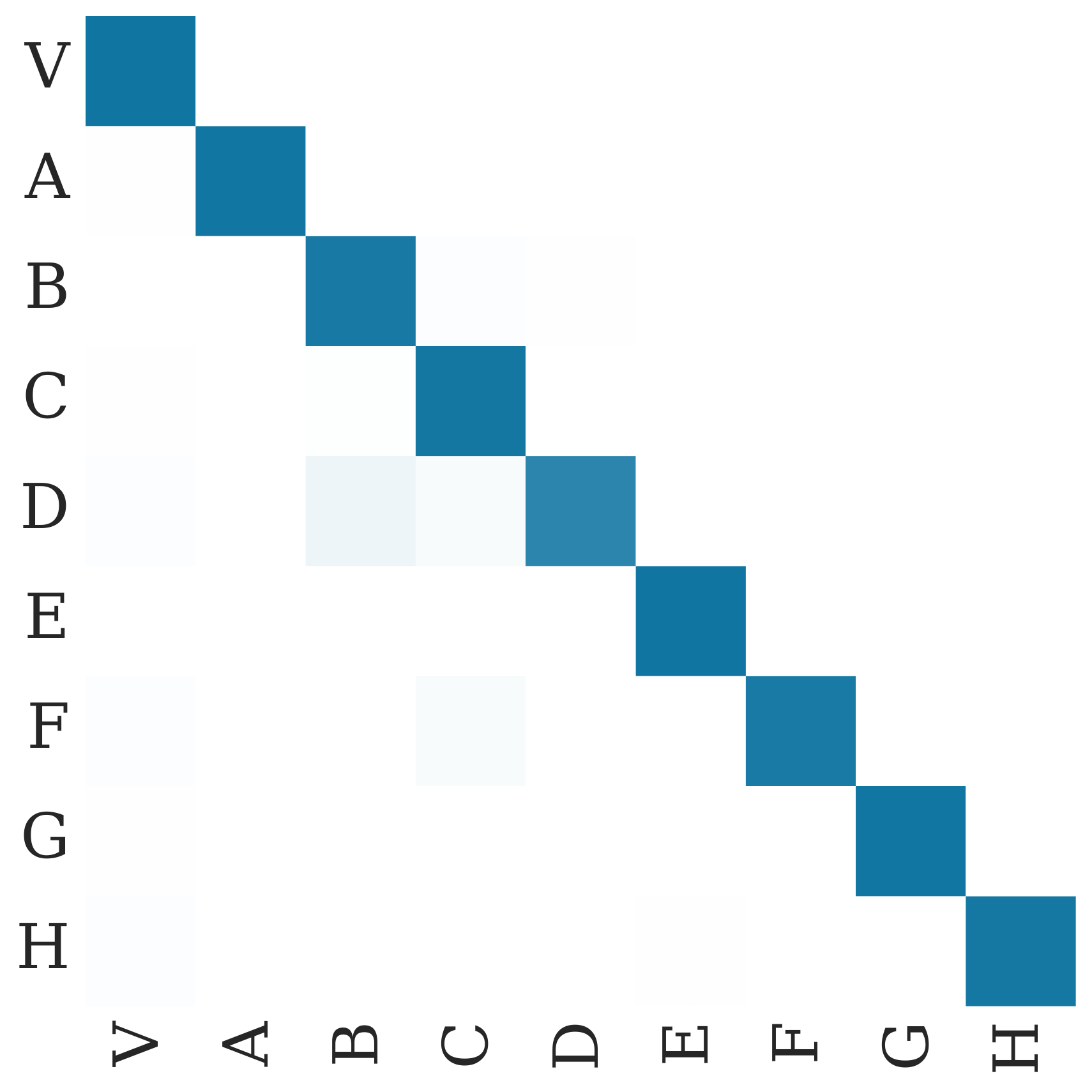}
        \caption{Viseme Categories}
    \end{subfigure}%
    \caption{Intra-viseme and inter-viseme confusion matrices, depicting the three categories with the most confusions, as well as the confusions between viseme clusters. Colours are row-normalised to emphasise the errors.}
    \label{fig:conf}
    \vspace{-1em}
\end{figure}
We compute confusion matrices between phonemes and then group phonemes into viseme clusters, following \cite{neti2000audio}. \Cref{fig:conf} shows the confusion matrices of the 3 most confused viseme categories, as well as the confusions between the viseme categories. The full phoneme confusion matrix is in \Cref{fig:conf_phoneme} in Appendix~\ref{app:phonemesvisemes}.

Given that the speakers are British, the confusion between /aa/ and /ay/ (\Cref{fig:conf}a) is most probably due to the fact that the first element, and the greater part, of the diphthong /ay/ is articulatorily identical with /aa/: an open back unrounded vowel \citep{ferragne2010formant}. 
The confusion of /ih/ (a rather close vowel) and /ae/ (a very open vowel) is at first glance surprising, but in fact in the sample /ae/ occurs only in the word \textit{at}, which is a function word normally pronounced with a reduced, weak vowel /ah/. /ah/ and /ih/ are the most frequent unstressed vowels and there is a good deal of variation within and between them, e.g. \textit{priv\underline{a}te} and \textit{watch\underline{e}s} \citep{cruttenden2014gimson}.

The confusion within the categories of bilabial stops /p b m/ and alveolar stops /t d n/ \mbox{(Figures~\ref{fig:conf}b-c)} is unsurprising: complete closure at the same place of articulation makes them look practically identical. The differences of velum action and vocal fold vibration are unobservable from the front.

Finally, the quality of the viseme categorisation of \cite{neti2000audio} is confirmed by the fact that the matrix in \Cref{fig:conf}d is diagonal, with only minor confusion between alveolar~(C) and palato-alveolar~(D) visemes. 
Articulatorily, alveolar /s z/ and palato-alveolar /sh zh/ fricatives are distinguished by only a small difference in tongue position: against the palate just behind the alveolar ridge, which is not easily observed from the front. The same can be said about dental /th/ and alveolar /t/.

\section{Conclusions}

We proposed LipNet, the first model to apply deep learning to end-to-end learning of a model that maps sequences of image frames of a speaker's mouth to entire sentences. The end-to-end model eliminates the need to segment videos into words before predicting a sentence. LipNet requires neither hand-engineered spatiotemporal visual features nor a separately-trained sequence model.

Our empirical evaluation illustrates the importance of spatiotemporal feature extraction and efficient temporal aggregation, confirming the intuition of \cite{easton1982perceptual}. Furthermore, LipNet greatly outperforms a human lipreading baseline, exhibiting $4.1\times$ better performance, and $4.8\%$ WER which is $2.8\times$ lower than the word-level state-of-the-art \citep{gergen2016dynamic} in the GRID corpus.

While LipNet is already an empirical success, the deep speech recognition literature \citep{amodei2015deep} suggests that performance will only improve with more data.
In future work, we hope to demonstrate this by applying LipNet to larger datasets, such as a sentence-level variant of that collected by \citet{chung2016lip}.

Some applications, such as silent dictation, demand the use of video only. However, to extend the range of potential applications of LipNet,  
we aim to apply this approach to a jointly trained audio-visual speech recognition model, where visual input assists with robustness in noisy environments.

\subsubsection*{Acknowledgments}

This work was supported by an Oxford-Google DeepMind Graduate Scholarship, the EPSRC, and CIFAR. We would also like to thank: NVIDIA for their generous donation of DGX-1 and GTX Titan X GPUs, used in our experiments; \'Aine Jackson, Brittany Klug and Samantha Pugh for helping us measure the experienced lipreader baseline; Mitko Sabev for his phonetics guidance; Odysseas Votsis for his video production help; and Alex Graves and Oiwi Parker Jones for helpful comments.

\renewcommand*{\bibfont}{\small}
\setlength{\bibsep}{1ex}
\bibliography{refs,liprefs}

\begin{thebibliography}{55}
\providecommand{\natexlab}[1]{#1}
\providecommand{\url}[1]{\texttt{#1}}
\expandafter\ifx\csname urlstyle\endcsname\relax
  \providecommand{\doi}[1]{doi: #1}\else
  \providecommand{\doi}{doi: \begingroup \urlstyle{rm}\Url}\fi

\bibitem[Almajai et~al.(2016)Almajai, Cox, Harvey, and Lan]{Almajai:2016}
I.~Almajai, S.~Cox, R.~Harvey, and Y.~Lan.
\newblock Improved speaker independent lip reading using speaker adaptive
  training and deep neural networks.
\newblock In \emph{IEEE International Conference on Acoustics, Speech and
  Signal Processing}, pp.\  2722--2726, 2016.

\bibitem[Amodei et~al.(2015)Amodei, Anubhai, Battenberg, Case, Casper,
  Catanzaro, Chen, Chrzanowski, Coates, Diamos, et~al.]{amodei2015deep}
D.~Amodei, R.~Anubhai, E.~Battenberg, C.~Case, J.~Casper, B.~Catanzaro,
  J.~Chen, M.~Chrzanowski, A.~Coates, G.~Diamos, et~al.
\newblock {Deep Speech 2: End-to-end speech recognition in English and
  Mandarin}.
\newblock \emph{arXiv preprint arXiv:1512.02595}, 2015.

\bibitem[Ashby(2013)]{ashby2013understanding}
P.~Ashby.
\newblock \emph{Understanding phonetics}.
\newblock Routledge, 2013.

\bibitem[Chung \& Zisserman(2016{\natexlab{a}})Chung and
  Zisserman]{chung2016lip}
J.~S. Chung and A.~Zisserman.
\newblock Lip reading in the wild.
\newblock In \emph{Asian Conference on Computer Vision}, 2016{\natexlab{a}}.

\bibitem[Chung \& Zisserman(2016{\natexlab{b}})Chung and
  Zisserman]{chung2016out}
J.~S. Chung and A.~Zisserman.
\newblock Out of time: automated lip sync in the wild.
\newblock In \emph{Workshop on Multi-view Lip-reading, ACCV},
  2016{\natexlab{b}}.

\bibitem[Chung et~al.(2014)Chung, Gulcehre, Cho, and
  Bengio]{chung2014empirical}
J.~Chung, C.~Gulcehre, K.~Cho, and Y.~Bengio.
\newblock Empirical evaluation of gated recurrent neural networks on sequence
  modeling.
\newblock \emph{arXiv preprint arXiv:1412.3555}, 2014.

\bibitem[Cooke et~al.(2006)Cooke, Barker, Cunningham, and Shao]{cooke2006audio}
M.~Cooke, J.~Barker, S.~Cunningham, and X.~Shao.
\newblock An audio-visual corpus for speech perception and automatic speech
  recognition.
\newblock \emph{The Journal of the Acoustical Society of America}, 120\penalty0
  (5):\penalty0 2421--2424, 2006.

\bibitem[Cruttenden(2014)]{cruttenden2014gimson}
A.~Cruttenden.
\newblock \emph{Gimson's pronunciation of English}.
\newblock Routledge, 2014.

\bibitem[Dahl et~al.(2012)Dahl, Yu, Deng, and Acero]{dahl2012context}
G.~E. Dahl, D.~Yu, L.~Deng, and A.~Acero.
\newblock Context-dependent pre-trained deep neural networks for
  large-vocabulary speech recognition.
\newblock \emph{IEEE Transactions on Audio, Speech, and Language Processing},
  20\penalty0 (1):\penalty0 30--42, 2012.

\bibitem[DeLand(1931)]{deland1931story}
F.~DeLand.
\newblock The story of lip-reading, its genesis and development.
\newblock 1931.

\bibitem[Easton \& Basala(1982)Easton and Basala]{easton1982perceptual}
R.~D. Easton and M.~Basala.
\newblock Perceptual dominance during lipreading.
\newblock \emph{Perception \& Psychophysics}, 32\penalty0 (6):\penalty0
  562--570, 1982.

\bibitem[Ferragne \& Pellegrino(2010)Ferragne and
  Pellegrino]{ferragne2010formant}
E.~Ferragne and F.~Pellegrino.
\newblock Formant frequencies of vowels in 13 accents of the british isles.
\newblock \emph{Journal of the International Phonetic Association}, 40\penalty0
  (01):\penalty0 1--34, 2010.

\bibitem[Fisher(1968)]{fisher1968confusions}
C.~G. Fisher.
\newblock Confusions among visually perceived consonants.
\newblock \emph{Journal of Speech, Language, and Hearing Research}, 11\penalty0
  (4):\penalty0 796--804, 1968.

\bibitem[Fu et~al.(2008)Fu, Yan, and Huang]{fu2008classification}
Y.~Fu, S.~Yan, and T.~S. Huang.
\newblock Classification and feature extraction by simplexization.
\newblock \emph{IEEE Transactions on Information Forensics and Security},
  3\penalty0 (1):\penalty0 91--100, 2008.

\bibitem[Garg et~al.(2016)Garg, Noyola, and Bagadia]{garglip}
A.~Garg, J.~Noyola, and S.~Bagadia.
\newblock Lip reading using {CNN} and {LSTM}.
\newblock Technical report, Stanford University, CS231n project report, 2016.

\bibitem[Gergen et~al.(2016)Gergen, Zeiler, Abdelaziz, Nickel, and
  Kolossa]{gergen2016dynamic}
S.~Gergen, S.~Zeiler, A.~H. Abdelaziz, R.~Nickel, and D.~Kolossa.
\newblock Dynamic stream weighting for turbo-decoding-based audiovisual {ASR}.
\newblock In \emph{Interspeech}, pp.\  2135--2139, 2016.

\bibitem[Goldschen et~al.(1997)Goldschen, Garcia, and
  Petajan]{goldschen1997continuous}
A.~J. Goldschen, O.~N. Garcia, and E.~D. Petajan.
\newblock Continuous automatic speech recognition by lipreading.
\newblock In \emph{Motion-Based recognition}, pp.\  321--343. Springer, 1997.

\bibitem[Graves \& Jaitly(2014)Graves and Jaitly]{graves2014towards}
A.~Graves and N.~Jaitly.
\newblock Towards end-to-end speech recognition with recurrent neural networks.
\newblock In \emph{International Conference on Machine Learning}, pp.\
  1764--1772, 2014.

\bibitem[Graves \& Schmidhuber(2005)Graves and
  Schmidhuber]{graves2005framewise}
A.~Graves and J.~Schmidhuber.
\newblock Framewise phoneme classification with bidirectional {LSTM} and other
  neural network architectures.
\newblock \emph{Neural Networks}, 18\penalty0 (5):\penalty0 602--610, 2005.

\bibitem[Graves et~al.(2006)Graves, Fern{\'a}ndez, Gomez, and
  Schmidhuber]{graves2006connectionist}
A.~Graves, S.~Fern{\'a}ndez, F.~Gomez, and J.~Schmidhuber.
\newblock Connectionist temporal classification: labelling unsegmented sequence
  data with recurrent neural networks.
\newblock In \emph{ICML}, pp.\  369--376, 2006.

\bibitem[Gurban \& Thiran(2009)Gurban and Thiran]{gurban2009information}
M.~Gurban and J.-P. Thiran.
\newblock Information theoretic feature extraction for audio-visual speech
  recognition.
\newblock \emph{IEEE Transactions on Signal Processing}, 57\penalty0
  (12):\penalty0 4765--4776, 2009.

\bibitem[He et~al.(2015)He, Zhang, Ren, and Sun]{he2015delving}
K.~He, X.~Zhang, S.~Ren, and J.~Sun.
\newblock Delving deep into rectifiers: Surpassing human-level performance on
  imagenet classification.
\newblock In \emph{IEEE International Conference on Computer Vision}, pp.\
  1026--1034, 2015.

\bibitem[Hilder et~al.(2009)Hilder, Harvey, and Theobald]{hilder2009comparison}
S.~Hilder, R.~Harvey, and B.-J. Theobald.
\newblock Comparison of human and machine-based lip-reading.
\newblock In \emph{AVSP}, pp.\  86--89, 2009.

\bibitem[Hinton et~al.(2012)Hinton, Deng, Yu, Dahl, Mohamed, Jaitly, Senior,
  Vanhoucke, Nguyen, Sainath, et~al.]{hinton2012deep}
G.~Hinton, L.~Deng, D.~Yu, G.~E. Dahl, A.-r. Mohamed, N.~Jaitly, A.~Senior,
  V.~Vanhoucke, P.~Nguyen, T.~N. Sainath, et~al.
\newblock Deep neural networks for acoustic modeling in speech recognition: The
  shared views of four research groups.
\newblock \emph{IEEE Signal Processing Magazine}, 29\penalty0 (6):\penalty0
  82--97, 2012.

\bibitem[Hochreiter \& Schmidhuber(1997)Hochreiter and
  Schmidhuber]{hochreiter1997long}
S.~Hochreiter and J.~Schmidhuber.
\newblock Long short-term memory.
\newblock \emph{Neural computation}, 9\penalty0 (8):\penalty0 1735--1780, 1997.

\bibitem[Hu et~al.(2016)Hu, Li, et~al.]{hu2016temporal}
D.~Hu, X.~Li, et~al.
\newblock Temporal multimodal learning in audiovisual speech recognition.
\newblock In \emph{IEEE Conference on Computer Vision and Pattern Recognition},
  pp.\  3574--3582, 2016.

\bibitem[Ji et~al.(2013)Ji, Xu, Yang, and Yu]{ji20133d}
S.~Ji, W.~Xu, M.~Yang, and K.~Yu.
\newblock 3d convolutional neural networks for human action recognition.
\newblock \emph{IEEE transactions on pattern analysis and machine
  intelligence}, 35\penalty0 (1):\penalty0 221--231, 2013.

\bibitem[Karpathy et~al.(2014)Karpathy, Toderici, Shetty, Leung, Sukthankar,
  and Fei-Fei]{karpathy2014large}
A.~Karpathy, G.~Toderici, S.~Shetty, T.~Leung, R.~Sukthankar, and L.~Fei-Fei.
\newblock Large-scale video classification with convolutional neural networks.
\newblock In \emph{Proceedings of the IEEE conference on Computer Vision and
  Pattern Recognition}, pp.\  1725--1732, 2014.

\bibitem[King(2009)]{king2009dlib}
D.~E. King.
\newblock Dlib-ml: A machine learning toolkit.
\newblock \emph{JMLR}, 10\penalty0 (Jul):\penalty0 1755--1758, 2009.

\bibitem[Kingma \& Ba(2014)Kingma and Ba]{kingma2014adam}
D.~Kingma and J.~Ba.
\newblock Adam: A method for stochastic optimization.
\newblock \emph{arXiv preprint arXiv:1412.6980}, 2014.

\bibitem[Koller et~al.(2015)Koller, Ney, and Bowden]{koller2015deep}
O.~Koller, H.~Ney, and R.~Bowden.
\newblock Deep learning of mouth shapes for sign language.
\newblock In \emph{ICCV Workshop on Assistive Computer Vision and Robotics},
  pp.\  85--91, 2015.

\bibitem[Krizhevsky et~al.(2012)Krizhevsky, Sutskever, and
  Hinton]{krizhevsky2012imagenet}
A.~Krizhevsky, I.~Sutskever, and G.~E. Hinton.
\newblock Imagenet classification with deep convolutional neural networks.
\newblock In \emph{Advances in neural information processing systems}, pp.\
  1097--1105, 2012.

\bibitem[Lucey \& Sridharan(2006)Lucey and Sridharan]{lucey2006patch}
P.~Lucey and S.~Sridharan.
\newblock Patch-based representation of visual speech.
\newblock In \emph{HCSNet workshop on use of vision in human-computer
  interaction}, pp.\  79--85, 2006.

\bibitem[Maas et~al.(2015)Maas, Xie, Jurafsky, and Ng]{maas2015lexicon}
A.~L. Maas, Z.~Xie, D.~Jurafsky, and A.~Y. Ng.
\newblock Lexicon-free conversational speech recognition with neural networks.
\newblock In \emph{NAACL}, 2015.

\bibitem[Matthews et~al.(2002)Matthews, Cootes, Bangham, Cox, and
  Harvey]{matthews2002extraction}
I.~Matthews, T.~F. Cootes, J.~A. Bangham, S.~Cox, and R.~Harvey.
\newblock Extraction of visual features for lipreading.
\newblock \emph{IEEE Transactions on Pattern Analysis and Machine
  Intelligence}, 24\penalty0 (2):\penalty0 198--213, 2002.

\bibitem[{McGurk} \& {MacDonald}(1976){McGurk} and {MacDonald}]{1976Nature}
H.~{McGurk} and J.~{MacDonald}.
\newblock {Hearing lips and seeing voices}.
\newblock \emph{Nature}, 264:\penalty0 746--748, 1976.

\bibitem[Neti et~al.(2000)Neti, Potamianos, Luettin, Matthews, Glotin, Vergyri,
  Sison, and Mashari]{neti2000audio}
C.~Neti, G.~Potamianos, J.~Luettin, I.~Matthews, H.~Glotin, D.~Vergyri,
  J.~Sison, and A.~Mashari.
\newblock Audio visual speech recognition.
\newblock Technical report, IDIAP, 2000.

\bibitem[Ngiam et~al.(2011)Ngiam, Khosla, Kim, Nam, Lee, and
  Ng]{ngiam2011multimodal}
J.~Ngiam, A.~Khosla, M.~Kim, J.~Nam, H.~Lee, and A.~Y. Ng.
\newblock Multimodal deep learning.
\newblock In \emph{International Conference on Machine Learning}, pp.\
  689--696, 2011.

\bibitem[Ninomiya et~al.(2015)Ninomiya, Kitaoka, Tamura, Iribe, and
  Takeda]{ninomiya2015integration}
H.~Ninomiya, N.~Kitaoka, S.~Tamura, Y.~Iribe, and K.~Takeda.
\newblock Integration of deep bottleneck features for audio-visual speech
  recognition.
\newblock In \emph{International Speech Communication Association}, 2015.

\bibitem[Noda et~al.(2014)Noda, Yamaguchi, Nakadai, Okuno, and
  Ogata]{noda2014lipreading}
K.~Noda, Y.~Yamaguchi, K.~Nakadai, H.~G. Okuno, and T.~Ogata.
\newblock Lipreading using convolutional neural network.
\newblock In \emph{INTERSPEECH}, pp.\  1149--1153, 2014.

\bibitem[Papandreou et~al.(2007)Papandreou, Katsamanis, Pitsikalis, and
  Maragos]{papandreou2007multimodal}
G.~Papandreou, A.~Katsamanis, V.~Pitsikalis, and P.~Maragos.
\newblock Multimodal fusion and learning with uncertain features applied to
  audiovisual speech recognition.
\newblock In \emph{Workshop on Multimedia Signal Processing}, pp.\  264--267,
  2007.

\bibitem[Papandreou et~al.(2009)Papandreou, Katsamanis, Pitsikalis, and
  Maragos]{papandreou2009adaptive}
G.~Papandreou, A.~Katsamanis, V.~Pitsikalis, and P.~Maragos.
\newblock Adaptive multimodal fusion by uncertainty compensation with
  application to audiovisual speech recognition.
\newblock \emph{IEEE Transactions on Audio, Speech, and Language Processing},
  17\penalty0 (3):\penalty0 423--435, 2009.

\bibitem[Petridis \& Pantic(2016)Petridis and Pantic]{petridis2016deep}
S.~Petridis and M.~Pantic.
\newblock Deep complementary bottleneck features for visual speech recognition.
\newblock In \emph{IEEE International Conference on Acoustics, Speech and
  Signal Processing (ICASSP)}, pp.\  2304--2308. IEEE, 2016.

\bibitem[Pitsikalis et~al.(2006)Pitsikalis, Katsamanis, Papandreou, and
  Maragos]{pitsikalis2006adaptive}
V.~Pitsikalis, A.~Katsamanis, G.~Papandreou, and P.~Maragos.
\newblock Adaptive multimodal fusion by uncertainty compensation.
\newblock In \emph{Interspeech}, 2006.

\bibitem[Potamianos et~al.(2003)Potamianos, Neti, Gravier, Garg, and
  Senior]{potamianos2003recent}
G.~Potamianos, C.~Neti, G.~Gravier, A.~Garg, and A.~W. Senior.
\newblock Recent advances in the automatic recognition of audiovisual speech.
\newblock \emph{Proceedings of the IEEE}, 91\penalty0 (9):\penalty0 1306--1326,
  2003.

\bibitem[Sagonas et~al.(2013)Sagonas, Tzimiropoulos, Zafeiriou, and
  Pantic]{sagonas2013300}
C.~Sagonas, G.~Tzimiropoulos, S.~Zafeiriou, and M.~Pantic.
\newblock 300 faces in-the-wild challenge: The first facial landmark
  localization challenge.
\newblock In \emph{IEEE International Conference on Computer Vision Workshops},
  pp.\  397--403, 2013.

\bibitem[Simonyan et~al.(2013)Simonyan, Vedaldi, and
  Zisserman]{simonyan2013deep}
K.~Simonyan, A.~Vedaldi, and A.~Zisserman.
\newblock Deep inside convolutional networks: Visualising image classification
  models and saliency maps.
\newblock \emph{arXiv preprint arXiv:1312.6034}, 2013.

\bibitem[Springenberg et~al.(2014)Springenberg, Dosovitskiy, Brox, and
  Riedmiller]{springenberg2014striving}
J.~T. Springenberg, A.~Dosovitskiy, T.~Brox, and M.~Riedmiller.
\newblock Striving for simplicity: The all convolutional net.
\newblock In \emph{ICLR Workshop}, 2014.

\bibitem[Sui et~al.(2015)Sui, Bennamoun, and Togneri]{sui2015listening}
C.~Sui, M.~Bennamoun, and R.~Togneri.
\newblock Listening with your eyes: Towards a practical visual speech
  recognition system using deep boltzmann machines.
\newblock In \emph{IEEE International Conference on Computer Vision}, pp.\
  154--162, 2015.

\bibitem[Takashima et~al.(2016)Takashima, Aihara, Takiguchi, Ariki, Mitani,
  Omori, and Nakazono]{takashima2016audio}
Y.~Takashima, R.~Aihara, T.~Takiguchi, Y.~Ariki, N.~Mitani, K.~Omori, and
  K.~Nakazono.
\newblock Audio-visual speech recognition using bimodal-trained bottleneck
  features for a person with severe hearing loss.
\newblock \emph{Interspeech}, pp.\  277--281, 2016.

\bibitem[Wand et~al.(2016)Wand, Koutnik, and Schmidhuber]{wand2016lipreading}
M.~Wand, J.~Koutnik, and J.~Schmidhuber.
\newblock Lipreading with long short-term memory.
\newblock In \emph{IEEE International Conference on Acoustics, Speech and
  Signal Processing}, pp.\  6115--6119, 2016.

\bibitem[Woodward \& Barber(1960)Woodward and Barber]{woodward1960phoneme}
M.~F. Woodward and C.~G. Barber.
\newblock Phoneme perception in lipreading.
\newblock \emph{Journal of Speech, Language, and Hearing Research}, 3\penalty0
  (3):\penalty0 212--222, 1960.

\bibitem[Zeiler \& Fergus(2014)Zeiler and Fergus]{zeiler2014visualizing}
M.~D. Zeiler and R.~Fergus.
\newblock Visualizing and understanding convolutional networks.
\newblock In \emph{European Conference on Computer Vision}, pp.\  818--833,
  2014.

\bibitem[Zhao et~al.(2009)Zhao, Barnard, and Pietikainen]{zhao2009lipreading}
G.~Zhao, M.~Barnard, and M.~Pietikainen.
\newblock Lipreading with local spatiotemporal descriptors.
\newblock \emph{IEEE Transactions on Multimedia}, 11\penalty0 (7):\penalty0
  1254--1265, 2009.

\bibitem[Zhou et~al.(2014)Zhou, Zhao, Hong, and
  Pietik{\"a}inen]{zhou2014review}
Z.~Zhou, G.~Zhao, X.~Hong, and M.~Pietik{\"a}inen.
\newblock A review of recent advances in visual speech decoding.
\newblock \emph{Image and Vision Computing}, 32\penalty0 (9):\penalty0
  590--605, 2014.

\end{thebibliography}
\bibliographystyle{include/iclr2017_conference}

\newpage
\appendix
\section{Architecture Details}
\label{app:hyperparams}

In this appendix, we provide additional details about the implementation and architecture.

\subsection{Implementation}

LipNet is implemented using Torch, the warp-ctc CTC library~\citep{amodei2015deep}, and Stanford-CTC's decoder implementation.
The network parameters were initialised using He initialisation \citep{he2015delving}, apart from the square GRU matrices that were orthogonally initialised, as described in \citep{chung2014empirical}. The models were trained with channel-wise dropout (dropout rate $p=0.5$) after each pooling layer and mini-batches of size $50$. We used the optimiser Adam \citep{kingma2014adam} with a learning rate of $10^{-4}$, and the default hyperparameters: a first-moment momentum coefficient of $0.9$, a second-moment momentum coefficient of $0.999$, and the numerical stability parameter $\epsilon=10^{-8}$.

The CER and WER scores were computed using CTC beam search with the following parameters for Stanford-CTC's decoder: beam width $200$, $\alpha=1$, and $\beta=1.5$. On top of that, we use a character 5-gram binarised language model, as suggested in \citep{graves2014towards}.

\subsection{LipNet Architecture}

 The videos were processed with DLib face detector \citep{king2009dlib} and the iBug face shape predictor with 68 landmarks \citep{sagonas2013300}.
 The RGB input frames were normalised using the following per-channel means and standard deviations: $[\mu_R=0.7136,\sigma_R=0.1138,\mu_G=0.4906,\sigma_G=0.1078,\mu_B=0.3283,\sigma_B=0.0917]$.

Table \ref{tbl:arch} summarises the LipNet architecture hyperparameters, where $T$ denotes time, $C$ denotes channels, $F$ denotes feature dimension, $H$ and $W$ denote height and width and $V$ denotes the number of words in the vocabulary including the CTC blank symbol.

\begin{table}[h]
\caption{LipNet architecture hyperparameters.}
\label{tbl:arch}
\begin{center}
\begin{tabular}{llll}
    \multicolumn{1}{c}{\bf Layer}  &\multicolumn{1}{c}{\bf Size / Stride / Pad }  &\multicolumn{1}{c}{\bf Input size} &\multicolumn{1}{l}{\bf Dimension order}
\\ \hline \\            \\
    STCNN  &  $3\times5\times5$ / $1,2,2$ / $1,2,2$  &  $75\ \, \times3\ \, \times50\times100$ & $T\times C\times H\times W$\\
    Pool  &  $1\times2\times2$ / $1,2,2$  &  $75\ \, \times32\times25\times50$ & $T\times C\times H\times W$ \\
    STCNN  &  $3\times5\times5$ / $1,2,2$ / $1,2,2$  &  $75\ \, \times32\times12\times25$ & $T\times C\times H\times W$ \\
    Pool  &  $1\times2\times2$ / $1,2,2$  &  $75\ \, \times64\times12\times25$ & $T\times C\times H\times W$ \\
    STCNN  &  $3\times3\times3$ / $1,2,2$ / $1,1,1$  &  $75\ \, \times64\times6\ \, \times12$ & $T\times C\times H\times W$ \\
    Pool  &  $1\times2\times2$ / $1,2,2$  &  $75\ \, \times96\times6\ \, \times12$ & $T\times C\times H\times W$ \\
    Bi-GRU  &  $256$   &  $75\times(96\times3\times6)$  & $T\times (C\times H\times W)$ \\
    Bi-GRU  &  $256$   &  $75\times512$  & $T\times F$ \\
    Linear  &  $27 + \text{blank}$  &  $75\times512$ & $T\times F$ \\
    Softmax  &   &  $75\times28$ & $T\times V$
\end{tabular} 
\end{center}
\end{table}

Note that spatiotemporal convolution sizes depend on the number of channels, and the kernel's three dimensions. Spatiotemporal kernel sizes are specified in the same order as the input size dimensions. The input dimension orderings are given in parentheses in the input size column.

Layers after the Bi-GRU are applied per-timestep.

\subsection{Baseline-LSTM Architecture}

Baseline-LSTM replicates the setup of \cite{wand2016lipreading}, and is trained the same way as LipNet. The model uses two LSTM layers with $128$ neurons. The input frames were converted to grayscale and were down-sampled to $50\times25$px, dropout $p=0$, and the parameters were initialised uniformly with values between~$[-0.05,0.05]$.

\section{Phonemes and Visemes}
\label{app:phonemesvisemes}

\begin{figure}[htb]
  \centering
  \includegraphics[width=0.8\linewidth]{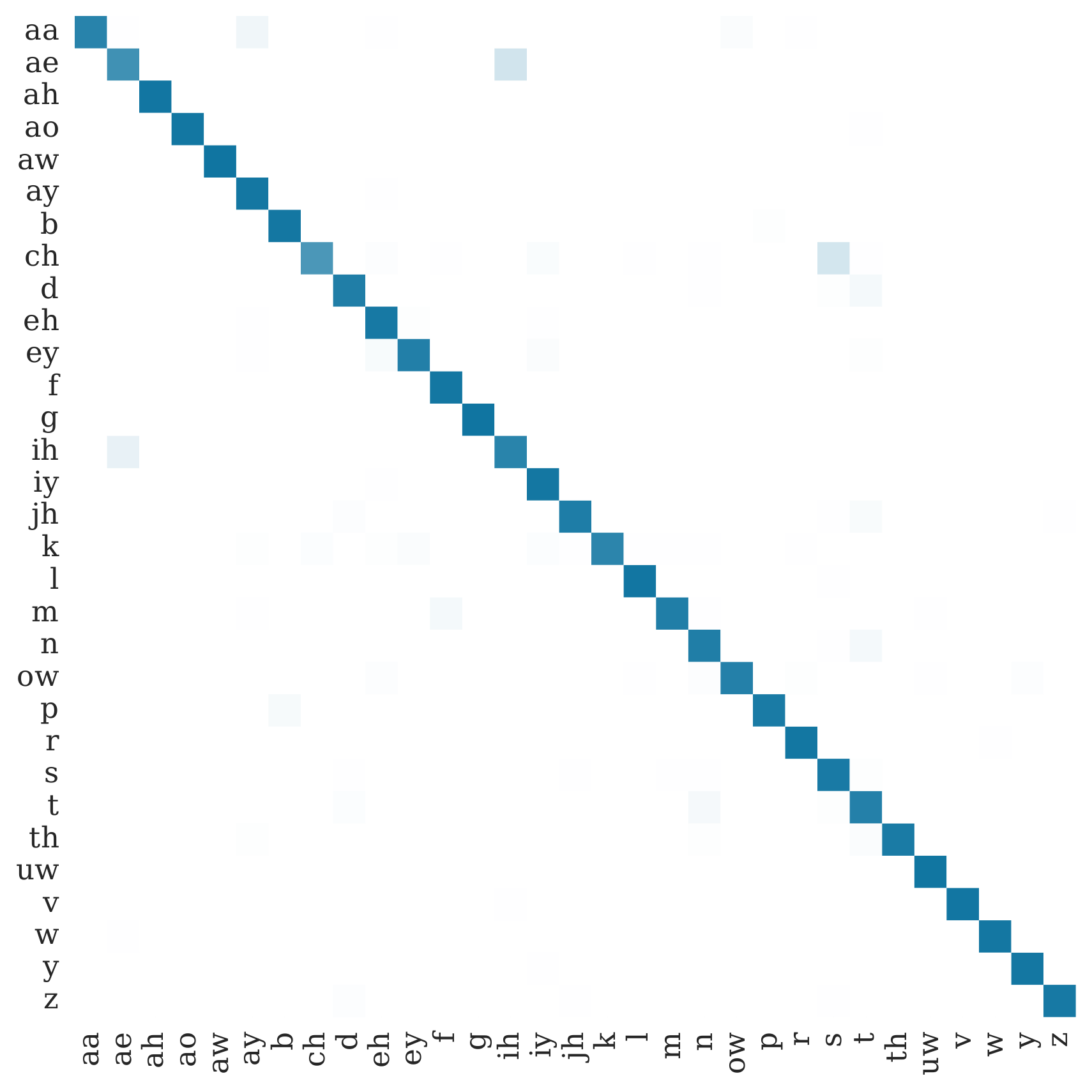}
  \caption{LipNet's full phoneme confusion matrix.}
  \label{fig:conf_phoneme}
\end{figure}

Table \ref{tbl:neti} shows the phoneme to viseme clustering of~\cite{neti2000audio} and Figure \ref{fig:conf_phoneme} shows LipNet's full phoneme confusion matrix.

\begin{table}[htb]
\caption{Phoneme to viseme clustering of~\cite{neti2000audio}.}
\label{tbl:neti}
\begin{center}
\begin{tabular}{lll}
\multicolumn{1}{c}{\bf Code}  &\multicolumn{1}{c}{\bf Viseme Class}  &\multicolumn{1}{c}{\bf Phonemes in Cluster}
\\ \hline \\
V1 & & /ao/ /ah/ /aa/ /er/ /oy/ /aw/ /hh/ \\
V2 & Lip-rounding based vowels & /uw/ /uh/ /ow/ \\
V3 & & /ae/ /eh/ /ey/ /ay/ \\
V4 & & /ih/ /iy/ /ax/ \\
A & Alveolar-semivowels & /l/ /el/ /r/ /y/ \\
B & Alveolar-fricatives & /s/ /z/ \\
C & Alveolar & /t/ /d/ /n/ /en/ \\
D & Palato-alveolar & /sh/ /zh/ /ch/ /jh/ \\
E & Bilabial & /p/ /b/ /m/ \\
F & Dental & /th/ /dh/ \\
G & Labio-dental & /f/ /v/ \\
H & Velar & /ng/ /k/ /g/ /w/ \\
S & Silence & /sil/ /sp/ 
\end{tabular} 
\end{center}
\end{table}

\end{document}